\documentclass[letterpaper, 10 pt, conference]{ieeeconf}  

\IEEEoverridecommandlockouts                              

\usepackage[font=small,labelfont=bf]{caption}
\usepackage[pdftex]{graphicx}
\usepackage{graphics} 
\usepackage{float} 
\usepackage{hyperref} 
\hypersetup{
  colorlinks   = true, 
  urlcolor     = blue, 
  linkcolor    = blue, 
}
\usepackage{epsfig} 
\usepackage{amsmath} 
\usepackage{amssymb}  
\usepackage{enumerate}
\usepackage{amsfonts}
\usepackage{subfig}
\usepackage{epstopdf}
\usepackage{booktabs}
\usepackage{array}
\usepackage{xcolor}
\usepackage[italian]{babel}
\usepackage[utf8]{inputenc}
\usepackage{verbatim}
\usepackage[active]{srcltx}
\usepackage{url}
\usepackage[T1]{fontenc}
\usepackage{mathtools} 
\usepackage{subfig}
\usepackage{cite}
\usepackage{microtype} 
\usepackage{bm}
\usepackage{dsfont}

\title{\LARGE \bf
Diff-MSM: Differentiable MusculoSkeletal Model for Simultaneous Identification of Human Muscle and Bone Parameters
}

\author{Yingfan Zhou$^{1\dag}$, Philip Sanderink$^{1\dag}$, Sigurd Jager Lemming$^{1}$, Cheng Fang$^{1*}$
\thanks{$^{1}$SDU Robotics, The Maersk Mc-Kinney Moller Institute, University of Southern Denmark.
$^\dag$These authors contributed equally to this work.}%
\thanks{This work is supported by the Innovation Fund Denmark Grand Solutions project, SENSIBLE (No. 2081-00031B), and the VPlegat project, MyoManager (No. 3410269). (\textit{$^*$Corresponding author: Cheng Fang, e-mail: chfa@mmmi.sdu.dk})}
}

\begin{document}

\maketitle
\thispagestyle{empty}
\pagestyle{empty}

\begin{abstract}

High-fidelity personalized human musculoskeletal models are crucial for simulating realistic behavior of physically coupled human-robot interactive systems and verifying their safety-critical applications in simulations before actual deployment, such as human-robot co-transportation and rehabilitation through robotic exoskeletons. Identifying subject-specific Hill-type muscle model parameters and bone dynamic parameters is essential for a personalized musculoskeletal model, but very challenging due to the difficulty of measuring the internal biomechanical variables \textit{in vivo} directly, especially the joint torques. In this paper, we propose using Differentiable MusculoSkeletal Model (Diff-MSM) to simultaneously identify its muscle and bone parameters with an end-to-end automatic differentiation technique differentiating from the measurable muscle activation, through the joint torque, to the resulting observable motion without the need to measure the internal joint torques. Through extensive comparative simulations, the results manifested that our proposed method significantly outperformed the state-of-the-art baseline methods, especially in terms of accurate estimation of the muscle parameters (i.e., initial guess sampled from a normal distribution with the mean being the ground truth and the standard deviation being 10\% of the ground truth could end up with an average of the percentage errors of the estimated values as low as 0.05\%). In addition to human musculoskeletal modeling and simulation, the new parameter identification technique with the Diff-MSM has great potential to enable new applications in muscle health monitoring, rehabilitation, and sports science.

\end{abstract}

\section{Introduction}

In the era of ubiquitous robots, various types of intelligent robots are engaged in our work and daily life. Physical human-robot interaction/collaboration is becoming an important form of human-robot coexistence for performing challenging tasks (e.g., human-robot co-manipulation), assist/augment human
performance (e.g., robotic exoskeleton), and study human sensorimotor control (e.g., human impedance regulation) \cite{fang2023human}. To make a physically coupled human-robot system safe, simulators which can simultaneously simulate both the robot dynamic and the human musculoskeletal models and test the system's interactive behavior are crucial before a robot controller can be deployed on a real system \cite{seth2018opensim, todorov2012mujoco, wang2022myosim, peternel2020human, fang2019a2ml, fang2012anthropomorphic}. Different people would respond to the same robot behavior differently. To capture this variance and achieve high-fidelity simulation results, personalizing a human musculoskeletal model (a plant from control theory perspective) and a human sensorimotor control model (a controller for human movement) for an individual is essential. This paper addresses the former, which refers to identification of the muscle and bone dynamic parameters of an individualized human musculoskeletal model.

In addition to realistic simulations, developing a convenient method to identify the parameters of a human musculoskeletal model accurately can also be helpful for muscle health monitoring and management, that is, we can potentially use the estimated muscle parameters to evaluate the muscle quality in terms of muscle strength and muscle power \cite{uchida2021biomechanics}. A handy muscle quality assessment method can impact widely and enable applications in biomechanics, rehabilitation, sports science, and medical diagnostics \cite{barbat2012assess}.

Most subject-specific musculoskeletal models are set up by scaling from a template model, where the default parameters are initially measured or estimated from the cadaver statistical data based on a particular population \cite{andersen2021introduction}. While accurately estimating the geometrical bone parameters, like a bone length, from the measurements of an optical marker-based motion capture system is relatively easy (which is currently the \textit{de facto} gold standard), directly measuring parameters related to bone inertial and muscle mechanical properties \textit{in vivo} is very challenging if not downright impossible. For instance, optimal muscle fiber length can be measured by invasive laser diffraction or microendoscopy, and muscle physiological cross-sectional area related to muscle maximum isometric force can be measured by magnetic resonance imaging. Because of these challenges, noninvasive and affordable methods of estimating these parameters through optimization are preferable to direct measurements. 

In this work, to our best knowledge, we propose simultaneously optimizing and estimating the subject-specific muscle and bone parameters of a musculoskeletal model for the first time. This research makes three primary contributions:

1) We develop a differentiable musculoskeletal model of the human arm (eight muscles, three movable bones, and five DoFs) that can be personalized for an individual. 

2) While differentiable musculoskeletal models have been used for musculoskeletal optimal control before \cite{michaud2022bioptim}, we propose using a Differentiable MusculoSkeletal Model (Diff-MSM) to simultaneously optimize and identify the subject-specific Hill-type muscle model parameters and the bone dynamic parameters for the first time.

3) We present extensive simulations that evaluate and compare the differentiable optimization against the state-of-the-art methods. The differentiable optimization can accurately identify model parameters and achieve significantly better results.

\begin{figure*} \centering
	{\includegraphics[width=2\columnwidth]{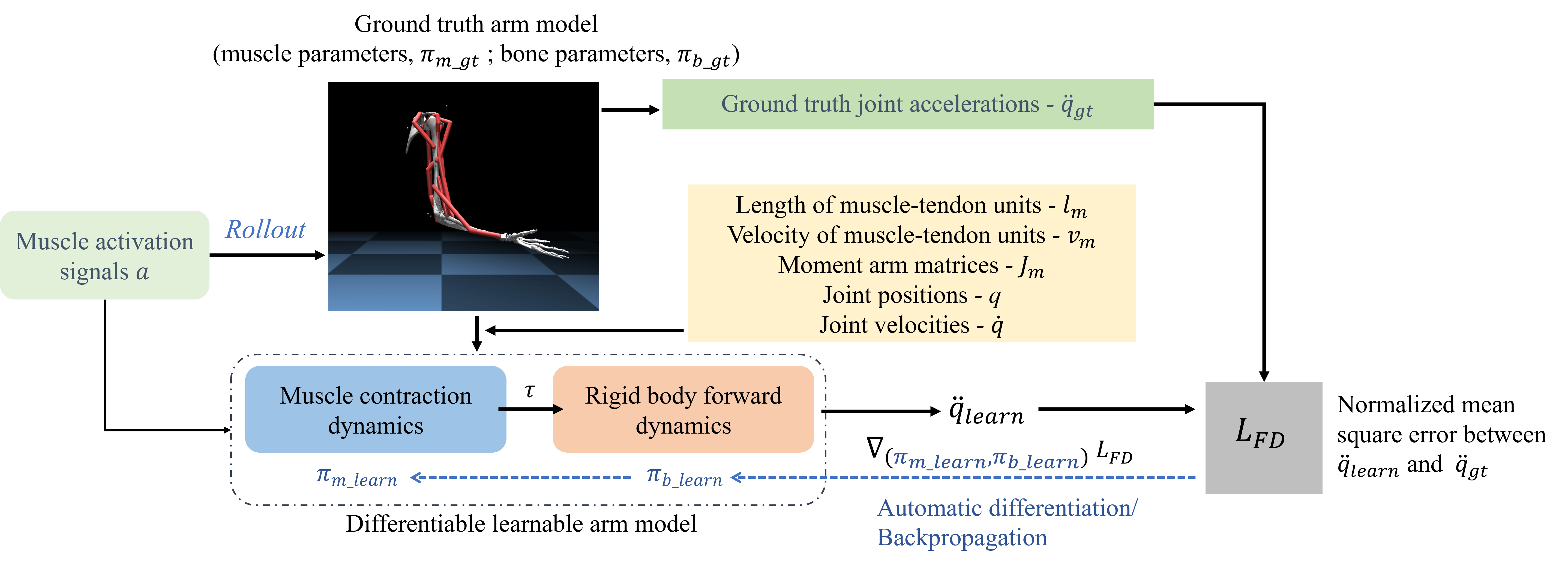} } 
	\vspace{-0.2cm}
	\caption{The pipeline of optimizing a differentiable musculoskeletal arm model's muscle and bone parameters with automatic differentiation.}
	\label{fig:pipeline}
	\vspace{-0.4cm}      
\end{figure*}

\section{Related works}

Dynamic parameter identification problem for multi-body systems, e.g., the identification of the mass, the center of mass, and the inertia tensor of a robot link, is a well-studied area in robotics \cite{lee2024robot,leboutet2021inertial}. By exploiting the property of linearity of the robot dynamic model concerning the dynamic parameters, the problem can be usually formulated as a linear least-squares estimation problem with measured robot motion and joint torque as input to solving this problem. However, the solutions cannot be guaranteed to be physically consistent, i.e., the mass has to be positive and the inertia tensor matrix has to be symmetric and positive-definite. To address this, the latest research focus in this field is how to formulate the problem as a semidefinite programming (SDP) optimization problem with linear matrix inequality constraints or a manifold optimization problem to efficiently and robustly find physically consistent solutions by considering possibly non-persistent excitation trajectories and noisy measurements \cite{sousa2014physical, wensing2017linear, lee2018geometric, traversaro2016identification}.  

Unfortunately, these mature techniques in robotics cannot be directly applied in the bone dynamic parameter estimation problem for the human skeleton because the internal joint torques of the human body cannot be directly measured. However, if we formulate the equation of the human full-body dynamics, and only use part of it, i.e., the floating base dynamics, to estimate the dynamic parameters of all the bones, measuring the joint torques can be avoided, and there are no internal forces and moments applied on the floating base of the human body. But, two force plates are usually needed to measure the external ground reaction forces and moments applied at the two feet, and the motion capture of the full-body movement is prone to having artifacts, which could lead to non-ideal parameter estimates \cite{jovic2016humanoid}. Alternatively, the subject-specific model parameters are often obtained from a template human musculoskeletal model by distributing the total body weight to the mass of individual bones following a mass distribution model based on statistical data \cite{winter2009biomechanics} and scaling the other dynamic parameters using scaling factors calculated by measured kinematic data \cite{andersen2021introduction, fang2018real}.

Once the bone dynamic parameters are estimated or obtained, the internal joint torques can be calculated using inverse dynamics of the skeleton with measured human body motion and possible external forces/moments as input. The calculated joint torques can be then used to estimate muscle parameters through muscle excitation and contraction dynamics by minimizing the errors between the calculated joint torques and those predicted by the muscle dynamics with neural excitations or muscle activations as input, which can be measured by surface EMG sensors, \cite{durandau2017robust, hayashibe2011muscle, venture2005identifying}. Since the linearity property of the skeleton does not remain in the highly nonlinear muscle dynamics, the error minimization problem is often solved by global optimization methods, such as the simulated annealing method \cite{goffe1994global}. However, the inaccuracy of the estimated or obtained bone parameters would greatly affect the accuracy of the calculated joint torques and hence the estimated muscle parameters as well \cite{rao2006influence}.

The aforementioned error minimization problem can be formulated as a nonlinear least-squares optimization, which is very common and at the heart of many problems in robotics. There is an emerging paradigm of solving this type of optimization problem by using differential optimization/simulator \cite{degrave2019differentiable, newbury2024review}, where the gradients of the error or loss relative to the model parameters, i.e., the optimization or learning parameters, are computed by automatic differentiation \cite{paszke2017automatic} on a computational graph reflecting the topological dependency between a large number of intermediate variables and parameters of a model, and gradient descent optimizers are used to minimize the loss to find the optimal model parameters. This new technique has been proved experimentally to be effective for system identification of different systems to solve the sim-2-real gap in recent works \cite{sutanto2020encoding, le2021differentiable, lutter2021differentiable, granados2022model, wang2023real2sim2real}. A big advantage of this differentiable optimization is that the computation graph connects many internal unmeasurable variables/parameters (e.g., joint torques, muscle and bone parameters in our case) to external observable variables (e.g., some muscle activations and body movement in our case), which can be utilized to infer and estimate the internal parameters without the need to measure the internal variables or separate the entire estimation process into multiple stages which was usually adopted in the previous studies, i.e., separation between bone dynamic parameter and muscle parameter estimation stages. In addition, another potential of this differentiable optimization is that the differentiable models can be inserted into and treated as individual layers in larger end-to-end trainable deep neural networks for a unified gradient-based optimization process. This will enable a trained hybrid model capable of dealing with complexity and generalizing to various scenarios by taking advantage of two complementary worlds \cite{amos2017optnet, heiden2020augmenting, wang2023pypose}. In this paper, we propose to use differential nonlinear optimization to simultaneously optimize and identify the muscle and bone parameters of the Diff-MSM.

\section{Our approach: DIff-MSM}

As shown in Fig. \ref{fig:pipeline}, a Diff-MSM is developed based on a human arm musculoskeletal model, MyoArm, from MyoSuite that is a simulation suite for musculoskeletal motor control powered by Mujoco \cite{todorov2012mujoco} as the physics engine. In our Diff-MSM, the 3-DoF shoulder and 2-DoF elbow joints are movable while the wrist and finger joints are locked. The differentiable model consists of eight muscles including the anterior (\textit{DELT}1) and posterior (\textit{DELT}3) portions of the Deltoid, the long (\textit{BIClong}) and short portions (\textit{BICshort}) of Biceps, the long (\textit{TRIlong}) and lateral (\textit{TRIlat}) portions of Triceps, Brachialis (\textit{BRA}) and Brachioradialis (\textit{BRD}), and three moveable bones including \textit{humerus}, \textit{ulna}, and \textit{radius}.

\subsection{Muscle Contraction Dynamics}

We adopt the Hill-type muscle model \cite{zajac1989muscle}\footnote{Elastic tendon and pennation angle at optimal fiber length modeling are currently not supported in the muscle model implemented in Mujoco, so the tendon is assumed to be inelastic and the penneation angle is not considered in our model.} to model the muscle contraction dynamics, namely, to calculate the muscle force and skeletal body joint torques from the muscle activation signals. By defining three predominant muscle dynamics parameters, which are optimal muscle fiber length \( l^{M}_{o}\), maximum isometric muscle force \( F^{M}_{o} \), and maximum muscle contraction velocity \( v^{M}_{max} \) in the model file, we can calculate the force produced by each muscle with
\begin{equation}\label{eq:contraction}
F_{m}=F_{active}(\tilde{l}_{m},\tilde{v}_{m},\pi_{m}, a)+ F_{passive}(\tilde{l}_{m},\pi_{m}),
\end{equation}
where \( F_{m} \) denotes the overall muscle force including the active muscle force \( F_{active} \) and the passive muscle force \( F_{passive} \). The active muscle force is calculated based on the force-length-velocity-activation relationship measured in plenty of experiments\cite{uchida2021biomechanics}. \( \tilde{l}_{m} \),  \( \tilde{v}_{m} \) and \( a \) represent the normalized length, normalized velocity of the muscle and the muscle activation signals respectively. \( \pi_{m} \) is a collection of muscles' parameters to be optimized with \(\pi_{mi} = [l^{M}_{o}, F^{M}_{o}, v^{M}_{max}] \in \mathbb{R}^{3}\) for each of the eight muscles.

\subsection{Forward Dynamics of the Rigid Skeletal Body}
After obtaining the muscle force generated by each muscle, with the moment arm matrix \( J_{m} \), in which each element is a moment arm of a muscle with respect to a joint that can be calculated with \(J_{m} = \frac{\partial l_{m}}{\partial q}\) \cite{zuo2024tsinghua}, the joint torques can be computed by
\begin{equation}\label{eq:torque}
    \tau=J_{m}^TF_{m}.
\end{equation}

Given the position \( q \), velocity \(\dot{q}\) and torque \( \tau \) of each joint at one simulation step, the joint accelerations \(\ddot{q}\) can be calculated with the \textit{Articulated Body Algorithm (ABA)} \cite{roy2014bodydyn} as a result of the skeleton forward dynamics as
\begin{equation}\label{eq:ABA}
    \ddot{q} = f_{ABA}({q}, \dot{q}, \tau, \pi_{b}),
\end{equation}
where \(\pi_{b} \) is a collection of the bones' parameters to be optimized with \(\pi_{bi} = [m, c_{x}, c_{y}, c_{z}, I_{xx}, I_{yy}, I_{zz}, I_{xy}, I_{xz}, I_{yz}]^T \in \mathbb{R}^{10}\) for each of three movable bones, where \( m \) is the bone mass, \( [c_{x}, c_{y}, c_{z}] \) are the coordinates of the center of mass and the last six parameters are the elements for making the inertia tensor $I_C$. 

\subsection{Loss Definition}
Based on the overall forward dynamics of the musculoskeletal system (i.e., the muscle contraction dynamics and the skeleton forward dynamics), the parameter identification problem is formulated as an optimization problem with the loss function defined as follows:
\begin{equation}\label{eq:loss}
    \ L_{FD} = \frac{\sum_{t=1}^{T} \left\| \ddot{q}_t - f_{FD} (q_t, \dot{q}_t, a, \pi) \right\|_2^2}{T\times var(\ddot{q}_t)},
\end{equation}
which is the normalized mean squared error between the joint accelerations calculated with the ground truth musculoskeletal model and those predicted by the learnable model with the same muscle activation signal trajectories as input. \( T \) represents the number of trajectory points in the dataset and  \( var(\ddot{q}_t)\) denotes the variance of the joint accelerations in the dataset.  \(f_{FD}\) is the overall forward dynamics function combining (\ref{eq:contraction})-(\ref{eq:ABA}). \(\pi\in\mathbb{R}^{54}\) represents the \(24(3\times8)\) muscle parameters and \(30(10\times3)\) bone parameters of the whole arm model to be optimized. Then we implement the entire musculoskeletal dynamics equation as a differentiable computation graph with PyTorch and exploit its embedded automatic differentiation tool \cite{paszke2017automatic} to calculate the loss gradient to optimize the 54 parameters simultaneously by minimizing the loss.

\subsection{Optimization} 

\textit{1) Parameter Constraints:} Since constraints cannot be explicitly added in the automatic differentiation, the constraints for guaranteeing the physical consistency of the optimized parameters are added implicitly by using the method proposed in\cite{sutanto2020encoding}, we represent the bone mass as: \(m = (\sqrt{\theta_{m}})^2 + b,\) where \(\theta_{m}\) is a new learnable mass parameter and \( b > 0 \) is a (non-learnable) small positive constant to ensure \( m > 0 \) The same tricks are implemented on \(l^{M}_{o}, F^{M}_{o}, v^{M}_{max}\) as well. For the inertia matrix $I_C$, it can be constructed as follow,
\begin{equation}
    \ I_C = Tr(\Sigma_C)I_{3 \times 3} - \Sigma_C
\end{equation}
with \( I_{3 \times 3} \) is a 3x3 identity matrix and \( Tr() \) is the matrix trace operation, then the inertia matrix will be positive definite and satisfy the triangular inequality constraint in the optimization process as long as \( \Sigma_C \succ 0 \) (a postive definite matrix), and it can be encoded implicitly by enforcing a Cholesky decomposition plus a small non-learnable positive bias \( c \) on the diagonal: 
\begin{equation}
\Sigma_C = L L^T + c I_{3 \times 3}, 
\end{equation}
where \( L \) is a lower triangular matrix as
\begin{equation}
    \ L = \begin{bmatrix}
I_{L_{{\Sigma_1}}} & 0 & 0 \\
I_{L_{{\Sigma_4}}} & I_{L_{{\Sigma_2}}} & 0 \\
I_{L_{{\Sigma_5}}} & I_{L_{{\Sigma_6}}} & I_{L_{{\Sigma_3}}}
\end{bmatrix}.
\end{equation}
Thus, \([I_{L_{{\Sigma_1}}}, I_{L_{{\Sigma_2}}}, I_{L_{{\Sigma_3}}}, I_{L_{{\Sigma_4}}}, I_{L_{{\Sigma_5}}}, I_{L_{{\Sigma_6}}}]\) can be used as the parameters to be optimized instead of the original inertia tensor parameters. Using these parametrization of $I_C$, no matter how the six new parameters change, $I_C$ would always lie on the constraint manifold. So, the original constrained optimization problem is converted to an unconstrained optimization problem which can be solved by the gradient based optimizers of PyTorch.

\textit{2) Learnable parameters:} Considering the effect of different parameter sizes/units on the sensitivity of the loss to the changes of different parameters, for instance, the size of muscle force (unit: \textit{Newton}) is typically between \(10^{2}\) and \(10^{3}\) whereas the size of the optimal muscle length (unit: \textit{meter}) is between \(10^{-2}\) and \(10^{-1}\). In order to increase loss sensitivity by encouraging small changes in the learnable parameters lead to larger loss changes, we utilized a strategy similar to what was proposed in \cite{rika2022diffcloud} to define a scalar multiplier for each of the 54 model parameters as a new learning parameter. This multiplier describes how much we would scale a model parameter from a default parameter value of a template model. We select \( [0.1, 10.0] \) as the range for each learnable multiplier. The relationship between the multiplier and the original parameter is
\begin{equation}\label{eq:sigmoid}
    \pi = \pi_{0} \times 10^{\text{sigmoid}(x) \times 2 - 1}
\end{equation}
where \( \pi \), \( \pi_{0} \) and \( x \) denote the original  parameter, the initial guess for the original parameter, and the learnable parameter, respectively. The sigmoid function is used in (\ref{eq:sigmoid}) to ensure that the final learnable parameter \( x \) is unconstrained which can be easily operated by gradient based optimizers while the defined range of the multiplier can be always respected through (\ref{eq:sigmoid}). 
\section{Comparative simulations}

In this section, we evaluate and compare the optimization performance of five methods with the same Diff-MSM:

\begin{itemize}
    \item M1: Simulated annealing to optimize only the muscle parameters $\pi_m$ given constant inaccurate bone dynamic parameters' estimates during the optimization (the state-of-the-art baseline method \cite{durandau2017robust});
    \item M2: Simulated Annealing to optimize the muscle and bone parameters, $\pi_m,\pi_b$, simultaneously;
    \item M3: Automatic differentiation for the gradient calculation and gradient descent with Adam optimizer \cite{kingma2014adam} to optimize only the muscle parameters, $\pi_m$, given the same inaccurate bone dynamic parameters as M1;
    \item M4: Automatic differentiation for the gradient calculation and gradient descent with Adam optimizer to optimize the muscle and skeletal parameters, $\pi_m,\pi_b$, simultaneously (proposed method);
    \item M5: Combination of M2 and M4 where Adam acts as the local optimizer for the simulated annealing method.
\end{itemize}

\subsection{Method description}

Simulated annealing is one of the global optimization methods that has been widely used in parameter identification tasks, for instance, in robot dynamic parameters identification \cite{gaz2019franka} as well as in human musculoskeletal dynamics parameter identification \cite{durandau2017robust}. We applied the simulated annealing algorithm in M1 and M2. In M1, only the muscle parameters were optimized. The bone parameters were perturbed with an error to emulate the parameter estimation inaccuracy caused by using a statistical mass distribution strategy. In M2, both muscle and bone parameters were optimized simultaneously to see whether optimizing both together would lead to better estimation accuracy. We used the SciPy Python package implementation of simulated annealing, as well as the PyTorch Minimize package \cite{gray2021minimize} to be able to optimize PyTorch variables directly. In M3, we used automatic differentiation with Adam to optimize muscle parameters while the bone parameters were perturbed in the same manner as in M1. In M4, we used automatic differentiation with Adam to optimize muscle and skeletal parameters simultaneously. In M5, simulated annealing was used to optimize muscle and skeletal parameters simultaneously, along with Adam as a local optimizer. The Scipy implementation of simulated annealing supports calling a customized local optimizer, whenever it finds a new best loss, or if there is no improvement for a number of iterations equal to the dimensionality of the search space, i.e., 36 in our case. We allowed Adam to run for a maximum of 100 iterations, or stopped it early if the loss became larger than the running average of the last ten losses. For the hyperparameters of the simulated annealing, the restart temperature ratio, initial temperature, and visit parameter were set to be 2e-15, 0.02, and 1.1, respectively. They were chosen this way to discourage exploration of the entire search space, as the initial guesses were relatively close to the ground truth parameters. For the Adam optimizer, a learning rate of 0.001 was used in relevant methods.

\begin{figure}[t]
    \centering
    \begin{minipage}{0.46\textwidth}
        \centering
        \includegraphics[width=\textwidth]{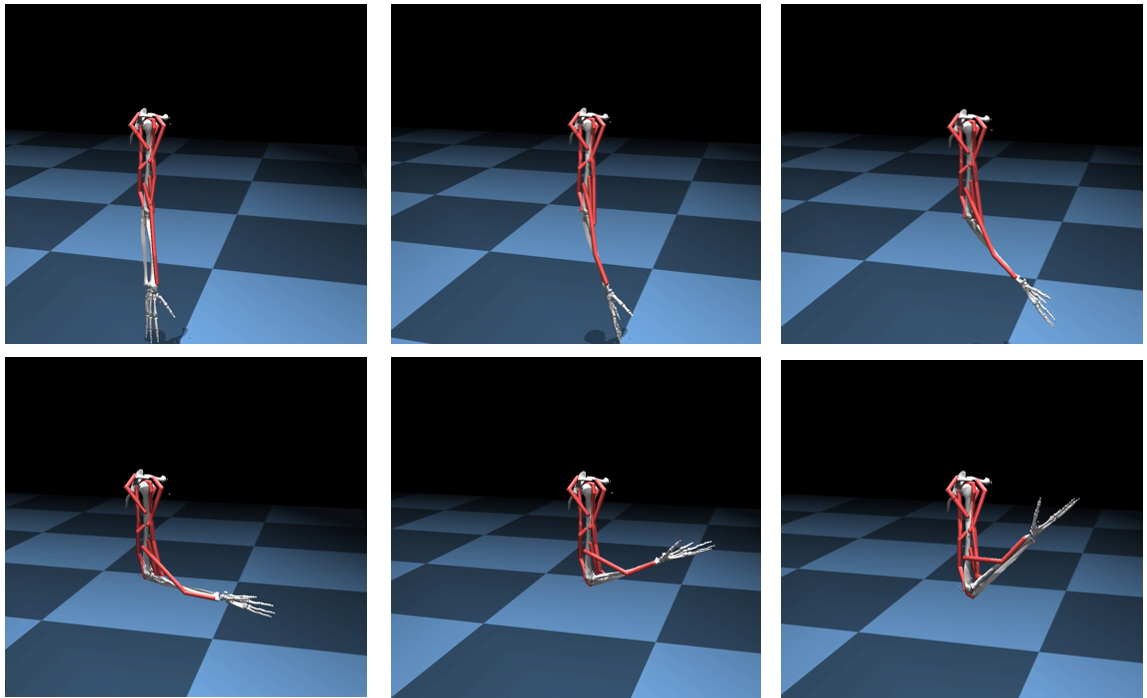}  
    \end{minipage}%
    \hfill
    \caption{MyoSuite snapshots of the arm posture taken at different times from a training dataset after eight muscles get actuated with continuous sinusoidal activation signals.}
    \label{fig:muscle}
    \vspace{-0.3cm}
\end{figure}

\subsection{Comparison settings}

\textit{1) Data Collection:} Before training, first we roll out the ground truth MyoArm model in simulation to generate the training dataset by activating all the muscles with continuous sinusoidal signal trajectories having different phases $\varphi$ with
\begin{equation}
    \ a(t) = A \sin(2 \pi f t + \varphi)+a_0
\end{equation}
As demonstrated in Fig. \ref{fig:pipeline}, the data collected at each simulation step include the length and velocity of the muscle, moment arm matrix, joint position, velocity and acceleration. Then, the same activation signals and the ground truth data except the ground truth joint accelerations will be input to the forward dynamics equations of the learnable model to calculate the joint accelerations and the corresponding loss at each time step. The entire training dataset contains 10K data points and is divided into 10 batches of 1K points in each training epoch. Fig. \ref{fig:muscle} shows a sequence of MyoSuite snapshots of the arm posture taken at six different times from the training dataset.

\textit{2) Parameter Initialization:} For each of the Diff-MSM parameters, we sample 50 different random initial points from a normal distribution,
\begin{equation}
    \pi_{0} \sim N(\pi_{gt}, 0.1\pi_{gt}),
    \label{eq:initvaluesample}
\end{equation}
where the mean is the ground truth value \( \pi_{gt}\) and the standard deviation is 10\% of the ground truth. The selection of the 10\% is based on the fact that the human body segment weight percentage (relative to the total body weight) does not vary significantly, especially among a certain population \cite{leva1996ajustments, winter2009biomechanics}. The same 50 initial points are used for all five methods. 

\subsection{Evaluation criteria}

To obtain a fair comparison between different optimization methods, three different criteria are proposed as follows. The first criterion \( C_{1}\) refers to the loss defined in (\ref{eq:loss}):
\begin{equation}
 C_{1}=L_{FD}.
\end{equation}

The second criterion \( C_{2}\) is defined as the distance between a vector of the estimated parameters and a vector of the ground truth parameters normalized by the magnitude of the ground truth vector, which is used to describe how far a solution is from the ground truth:
\begin{equation}
    C_{2}=\frac{\|\hat{\pi}-\pi_{gt}\|_2}{\|\pi_{gt}\|_2} \times 100\%,
\end{equation}
where ${\|\  \|_2}$ denotes the Eucleadian norm. 

The last criterion is the mean of the percentage errors of all the estimated parameters, which is also used to evaluate how close a solution is from the ground truth. However, the effect of the units of different parameters on \( C_{2}\) is considered and eliminated in \( C_{3}\), 
\begin{equation}
    C_{3}=\frac{1}{n_{p}}\sum_{i=1}^{n_{p}}\frac{|\hat{\pi}_i-\pi_{gt_{i}}|}{|\pi_{gt_{i}}|}\times 100\%,
\end{equation}
where \( n_{p}\) denotes the number of the estimated parameters, and ${|\  |}$ means the absolute value. In addition, we also use a special $C_3$ evaluation criterion, $C_3^m$, which refers to the mean of the percentage errors of only muscle parameters.

\vspace{-0.1cm}
\subsection{Results}

\begin{figure}[!t]
    \centering
    \includegraphics[trim = 20mm 0mm 0mm 5mm, clip, width=1.1\linewidth]{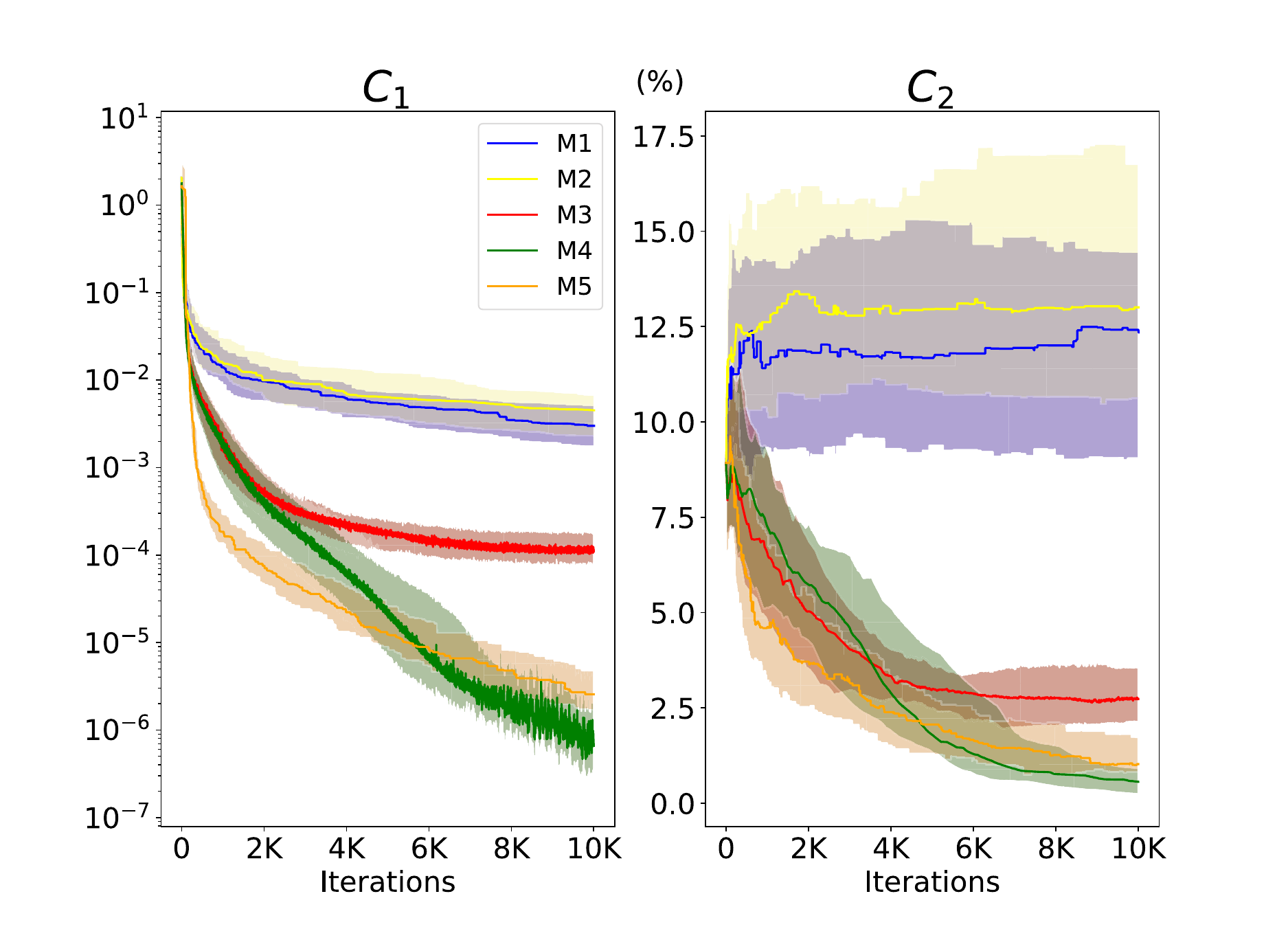}
    \caption{Loss curves in logarithmic scale ($C_1$) of 50 runs over 10K iterations of five different methods (M1: Simulated annealing with only muscle parameters; M2: Simulated annealing with all the parameters; M3: differential optimization with only muscle parameters; M4: differential optimization; M5: differentiable optimization and simulated annealing) (left). Distance error curves in percentage ($C_2$) of 50 runs over 10K iterations of M1-M5 (right). The solid line denotes the median while the upper and lower bounds of the shaded area represent upper and lower quartiles, respectively.}
    \label{fig:res:C1C2} 
\end{figure}
\begin{figure}[!t]
    \centering
    \includegraphics[trim = 20mm 0mm 0mm 5mm, clip, width=1.1\linewidth]{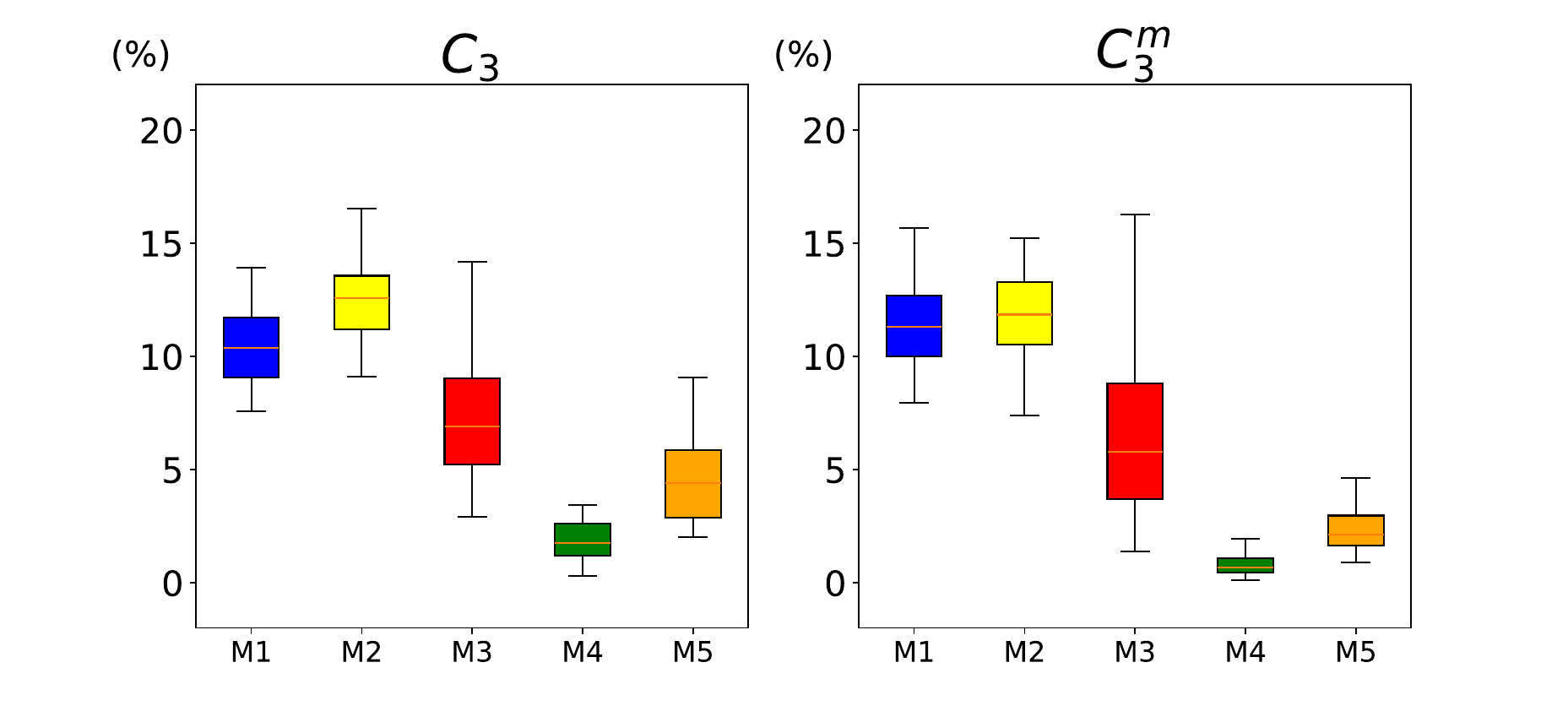}
\caption{Boxplot of the means of the percentage errors of all the estimated parameters ($C_3$) after 10K iterations of 50 runs for each of the five methods (M1: Simulated annealing with only muscle parameters; M2: Simulated annealing with all the parameters; M3: differential optimization with only muscle parameters; M4: differential optimization; M5: differentiable optimization and simulated annealing) (left), and boxplot of the means of the percentage errors of only estimated muscle parameters ($C_3^m$) after 10K iterations of 50 runs for each of M1-M5 (right). The original left boxplot contains outlier points beyond $1.5\times$IQR (Interquartile Range), which were omitted and not shown.}
    \label{fig:res:C3}
    \vspace{-0.3cm}
\end{figure}

50 runs were performed for each of the five methods over the same training dataset. In the comparative simulations, the six inertia tensor related parameters were not set as learning parameters for saving time, while they were included in the optimization process in the further simulations presented in Section \ref{sec:showoff}. All the simulations were run on a desktop equipped with a AMD Ryzen 5 7600X CPU and 32GB memory. The computation time was around $0.07$s per iteration for all methods, namely, the runtime for a single run for each method is around $11.7$ minutes. The results evaluated with the three criteria are shown in Fig. \ref{fig:res:C1C2} and \ref{fig:res:C3}. In Fig. \ref{fig:res:C1C2}, the solid line denotes the median of the criteria values of the 50 runs, while the upper and lower bounds of the shaded area represent upper and lower quartiles, representing how the criterion values are spread over iterations. Note that a logarithmic scale is applied in the left plot of Fig. \ref{fig:res:C1C2} to make their differences visually clear when comparing the loss curves.

It could be observed both from Fig. \ref{fig:res:C1C2} and Fig. \ref{fig:res:C3} that automatic differentiation related methods present much better results compared to simulated annealing based methods, i.e., M3 versus M1, and M4 versus M2, in terms of minimizing the loss and minimizing the distance error between the learnable parameters and the ground truth parameters. Importantly, simultaneous optimization of the muscle and bone parameters with automatic differentiation (M4) led to better results compared to the optimization of only the muscle parameters (M3), which was not observed when simulated annealing was used (M1 versus M2). Surprisingly, combining differentiable optimization and simulated annealing (M5) did not bring better results compared to pure differentiable optimization (M4). Furthermore, Fig. \ref{fig:res:C3} illustrates that, in M4 and M5, a larger portion of the estimation error seems to lie in the bone parameters $m,c_x,c_y,c_y$, as removing them from the error calculation decreases the median of the estimation errors ($C_3^m$ versus $C_3$ in the two subplots), which seems to imply that the muscle parameters can be estimated more accurately compared to the bone parameters. The comparative results manifest that our proposed method, M4, outperforms the other methods in terms of all the criteria.

\begin{figure*}[!t] \centering
	{\includegraphics[width=2\columnwidth]{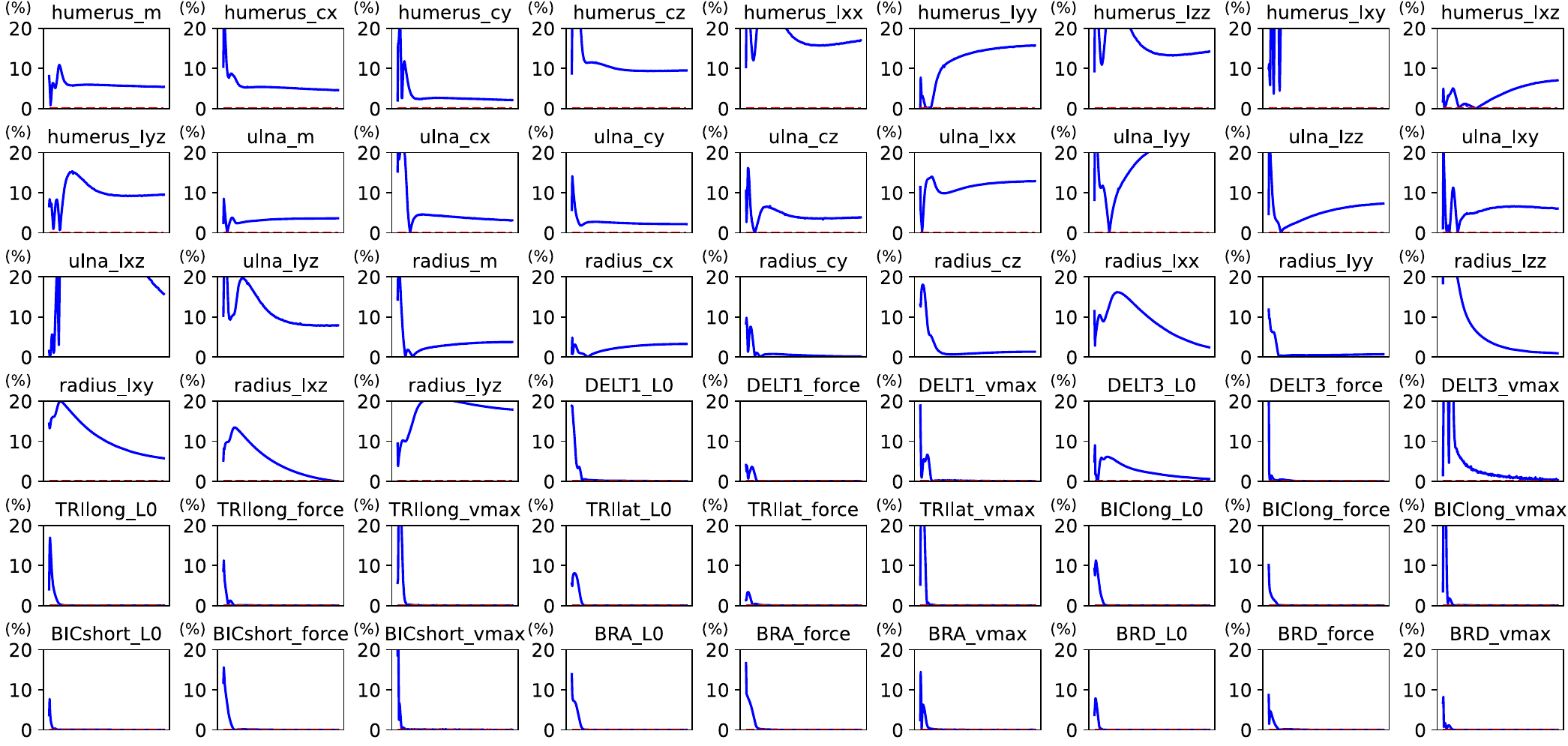} } 
	\vspace{-0.2cm}
	\caption{Percentage errors of all the estimated musculoskeletal parameters (including inertia tensor related parameters) over 100k iterations of the best run (out of eight runs) with the differentiable optimization, M4.}
	\label{fig:parameters evolution}
	\vspace{-0.1cm}      
\end{figure*}
\begin{figure*}[!t] 
    \centering
	{\includegraphics[width=2\columnwidth]{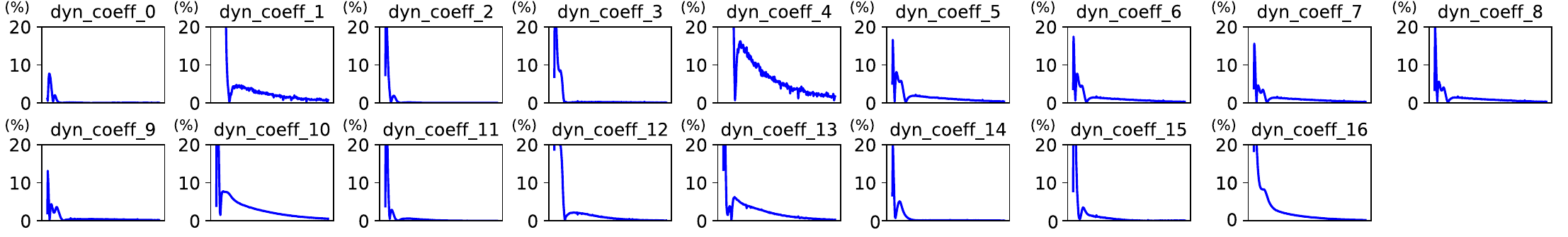} } 
	\vspace{-0.2cm}
        \captionsetup{justification=raggedright,singlelinecheck=false}
	\caption{Percentage errors of the corresponding dynamic coefficients of the estimated bone parameters in Fig. \ref{fig:parameters evolution}. }
	\label{fig:parameters evolution2}
	\vspace{-0.4cm}      
\end{figure*}
\begin{figure}[!t]
     \centering
     \includegraphics[trim = 0mm 0mm 0mm 0mm, clip, width=1\linewidth]{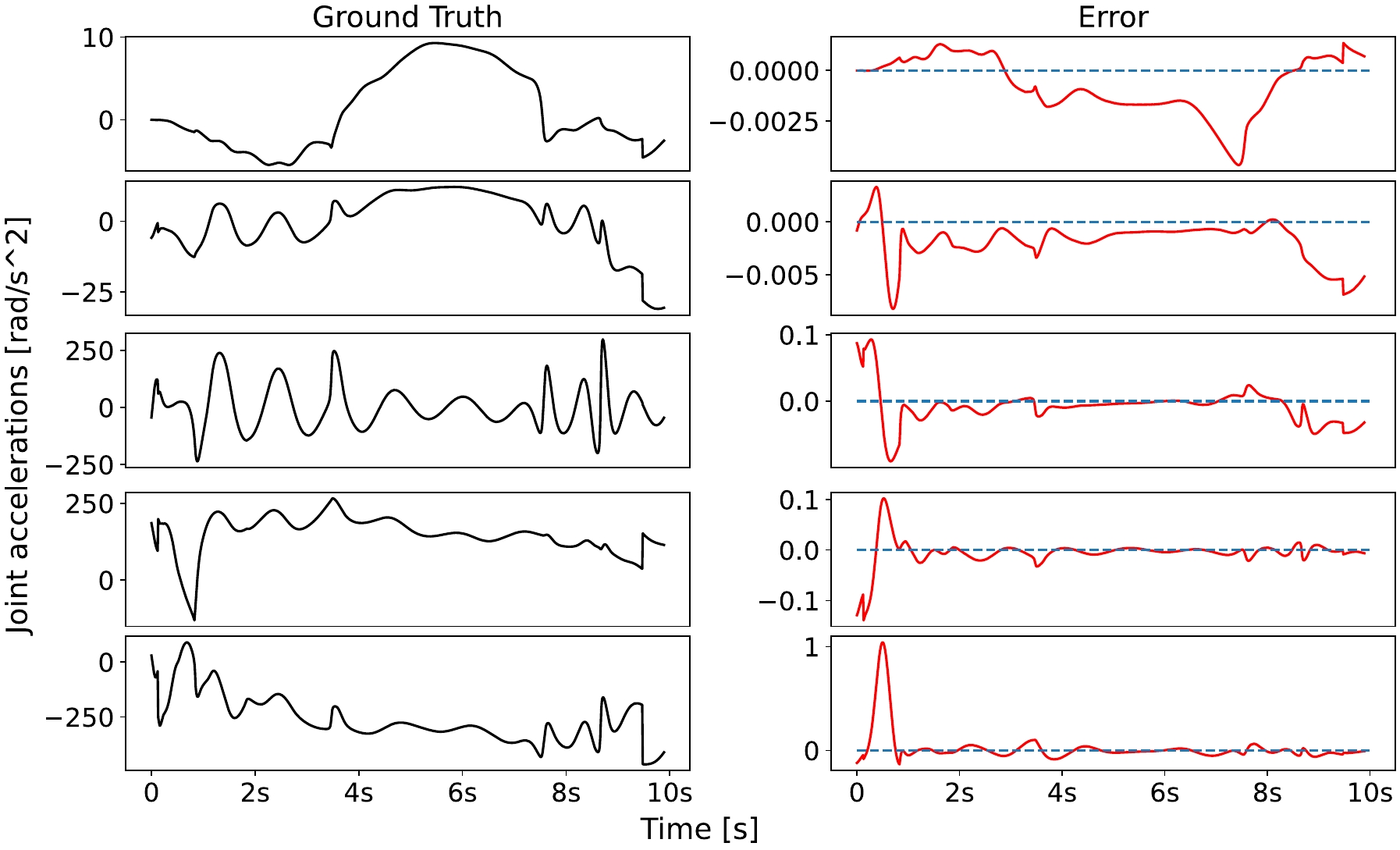}
     \caption{Ground truth joint accelerations of the five joints of the arm Diff-MSM from a testing data (left). Prediction errors of the five joint accelerations using the best estimated muscle and bone parameters of the arm Diff-MSM (right).}
     \label{fig:resBigTesting}
\end{figure}

\vspace{-0.1cm}
\subsection{Further simulations with differentiable optimization}\label{sec:showoff}
To test the performance of the M4 method further, we included the six inertia matrix parameters back as the learnable parameters and performed eight runs of simulations with 10k epochs per run. Then, the best run was picked based on the $C_3^m$, and its estimated parameters are shown in Fig. \ref{fig:parameters evolution}. It can be observed that all the muscle parameters were estimated extremely well, which steadily converged to their ground truth values as the epoch increased. Although most of the bone parameters did not converge to their ground truth values, the $C_1$, $C_2$, $C_3$, and $C_3^m$ of the estimated parameters after the last iteration achieved 6.9e-9, 0.04\%, 4.66\%, and 0.05\%, respectively, which indicates a really good solution was found. There exists redundancy in the bone dynamic parameters (including the inertia matrix parameters), i.e., an infinite number of parameter solutions can achieve the same dynamic behavior of bones. However, a set of dynamic coefficients, as linear combinations of the original dynamic parameters, can be uniquely identified \cite{gaz2019dynamic}. These coefficients are a minimum set that can determine the dynamic behavior of bones. Therefore, the original 30 bone dynamic parameters were converted to a set of 17 dynamic coefficients using the approach in \cite{atkeson1986estimation}, and the convergence of these coefficients to their ground truth values can be seen in Fig. \ref{fig:parameters evolution2}.


To further validate the generalization performance of the best solution of the identified parameters from Fig. \ref{fig:parameters evolution}, we generated a new testing dataset from Mujoco, with which we calculated the joint accelerations using the best found parameters. The errors between these calculated five joint accelerations and their ground truth values are shown in Fig. \ref{fig:resBigTesting}, which were relatively small compared to their ground truth values shown in the same figure as well. The normalized mean squared error on the testing dataset is 7.75e-8 ($C_1$). Thus, the best estimated musculoskeletal parameters demonstrated great generalization ability to the new dataset.

\vspace{-0.1cm}
\section{Implications and Discussion}

From the right subplot of Fig. \ref{fig:res:C1C2}, it was observed that the differentiable nonlinear optimization alone M4 significantly outperformed the state-of-the art methods, M1, in terms of the accuracy of the muscle parameters' estimates, i.e., the optimal muscle fiber length, the maximum isometric force, and the maximum contraction velocity. These muscle parameters can be used to evaluate the muscle strength and power \cite{uchida2021biomechanics} as two important criteria of the health status of an individual muscle. Currently, the muscle strength and power are usually measured by isokinetic dynamometers \cite{barbat2012assess} only for a group of muscles, e.g., the group of muscles contributing to the torque generation at the knee joint, rather than for individual muscles. This muscle parameter identification technique can enable monitoring and fine management of our muscle health.

Once the muscle parameters are accurately estimated, muscle contraction dynamics can be used to calculate good estimates of the joint torques that can then be used to improve the estimation of the bone dynamic parameters/coefficients. For instance, we could use the state-of-the-art estimation methods of robot dynamic parameters to formulate the bone dynamic parameter estimation problem as a SDP problem, which guarantees a physically consistent and global minimum solution and can be efficiently and reliably solved by existing available solvers.

With our proposed method (M4), all the muscle parameters and bone dynamic coefficients can converge to their ground truth values when the initial guess is not far from the ground truth (they are sampled from a normal distribution with a standard deviation being 10\% of the ground truth). It appears that our optimization problem exhibits fairly good local convexity of the loss landscape around the ground truth parameters, which facilitates the local optimizer, Adam. Whether or not there is an indeterminacy of the muscle parameter (and the bone dynamic coefficient) solution, similar to the relationship between bone dynamic parameters and coefficients shown in Fig. \ref{fig:parameters evolution} and \ref{fig:parameters evolution2}, needs further investigation with more initial guesses deviated from the ground truth significantly. Nevertheless, the muscle parameters usually do not differ considerably from those of a scaled template musculoskeletal model, especially among a particular healthy population.

\vspace{-0.1cm}
\section{Conclusions and Future Work}

In this pilot simulation study, we developed a differentiable human musculoskeletal arm model for simultaneous identification of the Hill-type muscle parameters and the bone dynamic parameters for an individual's arm. Automatic optimization was used to calculate the loss gradient with respect to all the model parameters, and gradient descent with Adam optimizer was adopted to find the optimal parameters by minimizing the loss. By comparing with the state-of-the-art baseline methods, we found our proposed method outperformed the other methods, especially for the accurate estimation of muscle parameters, which would impact widely in applications of muscle health monitoring, rehabilitation, and sports science.

In the future, we will improve the bone parameter estimation by leveraging the accurate joint torque estimates enabled by the precise muscle parameter estimates from this work with the state-of-the-art robot parameter estimation methods. Moreover, it is interesting to investigate the hypothesized local convexity around the ground truth and test if the local convexity also exists in other human musculoskeletal models. In addition, real experiments will be carried out to validate the effectiveness of the proposed method with real human subjects. To achieve this, we plan to design a neural network that maps measured human movement to muscle activations, including those of the deeper muscles that cannot be directly measured. In this way, the Diff-MSM can be combined with this neural network and trained together as a whole.



\bibliographystyle{IEEEtran}	
\bibliography{IEEEabrv,references}

\begin{thebibliography}{10}
\providecommand{\url}[1]{#1}
\csname url@samestyle\endcsname
\providecommand{\newblock}{\relax}
\providecommand{\bibinfo}[2]{#2}
\providecommand{\BIBentrySTDinterwordspacing}{\spaceskip=0pt\relax}
\providecommand{\BIBentryALTinterwordstretchfactor}{4}
\providecommand{\BIBentryALTinterwordspacing}{\spaceskip=\fontdimen2\font plus
\BIBentryALTinterwordstretchfactor\fontdimen3\font minus
  \fontdimen4\font\relax}
\providecommand{\BIBforeignlanguage}[2]{{%
\expandafter\ifx\csname l@#1\endcsname\relax
\typeout{** WARNING: IEEEtran.bst: No hyphenation pattern has been}%
\typeout{** loaded for the language `#1'. Using the pattern for}%
\typeout{** the default language instead.}%
\else
\language=\csname l@#1\endcsname
\fi
#2}}
\providecommand{\BIBdecl}{\relax}
\BIBdecl

\bibitem{fang2023human}
C.~Fang, L.~Peternel, A.~Seth, M.~Sartori, K.~Mombaur, and E.~Yoshida, ``Human
  modeling in physical human-robot interaction: A brief survey,'' \emph{IEEE
  Robotics and Automation Letters}, vol.~8, no.~9, pp. 5799--5806, 2023.

\bibitem{seth2018opensim}
A.~Seth, J.~L. Hicks, T.~K. Uchida, A.~Habib, C.~L. Dembia, J.~J. Dunne, C.~F.
  Ong, M.~S. DeMers, A.~Rajagopal, M.~Millard \emph{et~al.}, ``Opensim:
  Simulating musculoskeletal dynamics and neuromuscular control to study human
  and animal movement,'' \emph{PLoS computational biology}, vol.~14, no.~7, p.
  e1006223, 2018.

\bibitem{todorov2012mujoco}
E.~Todorov, T.~Erez, and Y.~Tassa, ``Mujoco: A physics engine for model-based
  control,'' in \emph{2012 IEEE/RSJ international conference on intelligent
  robots and systems}.\hskip 1em plus 0.5em minus 0.4em\relax IEEE, 2012, pp.
  5026--5033.

\bibitem{wang2022myosim}
H.~Wang, V.~Caggiano, G.~Durandau, M.~Sartori, and V.~Kumar, ``Myosim: Fast and
  physiologically realistic mujoco models for musculoskeletal and exoskeletal
  studies,'' in \emph{2022 International Conference on Robotics and Automation
  (ICRA)}.\hskip 1em plus 0.5em minus 0.4em\relax IEEE, 2022, pp. 8104--8111.

\bibitem{peternel2020human}
L.~Peternel, C.~Fang, M.~Laghi, A.~Bicchi, N.~Tsagarakis, and A.~Ajoudani,
  ``Human arm posture optimisation in bilateral teleoperation through interface
  reconfiguration,'' in \emph{2020 8th IEEE RAS/EMBS International Conference
  for Biomedical Robotics and Biomechatronics (BioRob)}.\hskip 1em plus 0.5em
  minus 0.4em\relax IEEE, 2020, pp. 1102--1108.

\bibitem{fang2019a2ml}
C.~Fang, X.~Ding, C.~Zhou, and N.~Tsagarakis, ``A2ml: A general human-inspired
  motion language for anthropomorphic arms based on movement primitives,''
  \emph{Robotics and Autonomous Systems}, vol. 111, pp. 145--161, 2019.

\bibitem{fang2012anthropomorphic}
C.~Fang and X.~Ding, ``Anthropomorphic arm kinematics oriented to movement
  primitive of human arm triangle,'' \emph{Jiqiren/Robot}, 2012.

\bibitem{uchida2021biomechanics}
T.~K. Uchida and S.~L. Delp, \emph{Biomechanics of movement: the science of
  sports, robotics, and rehabilitation}.\hskip 1em plus 0.5em minus 0.4em\relax
  Mit Press, 2021.

\bibitem{barbat2012assess}
S.~Barbat-Artigas, Y.~Rolland, M.~Zamboni, and M.~Aubertin-Leheudre, ``How to
  assess functional status: a new muscle quality index,'' \emph{The Journal of
  nutrition, health and aging}, vol.~16, no.~1, pp. 67--77, 2012.

\bibitem{andersen2021introduction}
M.~S. Andersen, ``Introduction to musculoskeletal modelling,'' in
  \emph{Computational modelling of biomechanics and biotribology in the
  musculoskeletal system}.\hskip 1em plus 0.5em minus 0.4em\relax Elsevier,
  2021, pp. 41--80.

\bibitem{michaud2022bioptim}
B.~Michaud, F.~Bailly, E.~Charbonneau, A.~Ceglia, L.~Sanchez, and M.~Begon,
  ``Bioptim, a python framework for musculoskeletal optimal control in
  biomechanics,'' \emph{IEEE Transactions on Systems, Man, and Cybernetics:
  Systems}, vol.~53, no.~1, pp. 321--332, 2022.

\bibitem{lee2024robot}
T.~Lee, J.~Kwon, P.~M. Wensing, and F.~C. Park, ``Robot model identification
  and learning: A modern perspective,'' \emph{Annual Review of Control,
  Robotics, and Autonomous Systems}, vol.~7, 2024.

\bibitem{leboutet2021inertial}
Q.~Leboutet, J.~Roux, A.~Janot, J.~R. Guadarrama-Olvera, and G.~Cheng,
  ``Inertial parameter identification in robotics: A survey,'' \emph{Applied
  Sciences}, vol.~11, no.~9, p. 4303, 2021.

\bibitem{sousa2014physical}
C.~D. Sousa and R.~Cortesao, ``Physical feasibility of robot base inertial
  parameter identification: A linear matrix inequality approach,'' \emph{The
  International Journal of Robotics Research}, vol.~33, no.~6, pp. 931--944,
  2014.

\bibitem{wensing2017linear}
P.~M. Wensing, S.~Kim, and J.-J.~E. Slotine, ``Linear matrix inequalities for
  physically consistent inertial parameter identification: A statistical
  perspective on the mass distribution,'' \emph{IEEE Robotics and Automation
  Letters}, vol.~3, no.~1, pp. 60--67, 2017.

\bibitem{lee2018geometric}
T.~Lee and F.~C. Park, ``A geometric algorithm for robust multibody inertial
  parameter identification,'' \emph{IEEE Robotics and Automation Letters},
  vol.~3, no.~3, pp. 2455--2462, 2018.

\bibitem{traversaro2016identification}
S.~Traversaro, S.~Brossette, A.~Escande, and F.~Nori, ``Identification of fully
  physical consistent inertial parameters using optimization on manifolds,'' in
  \emph{2016 IEEE/RSJ International Conference on Intelligent Robots and
  Systems (IROS)}.\hskip 1em plus 0.5em minus 0.4em\relax IEEE, 2016, pp.
  5446--5451.

\bibitem{jovic2016humanoid}
J.~Jovic, A.~Escande, K.~Ayusawa, E.~Yoshida, A.~Kheddar, and G.~Venture,
  ``Humanoid and human inertia parameter identification using hierarchical
  optimization,'' \emph{IEEE Transactions on Robotics}, vol.~32, no.~3, pp.
  726--735, 2016.

\bibitem{winter2009biomechanics}
D.~A. Winter, \emph{Biomechanics and motor control of human movement}.\hskip
  1em plus 0.5em minus 0.4em\relax John wiley \& sons, 2009.

\bibitem{fang2018real}
C.~Fang, A.~Ajoudani, A.~Bicchi, and N.~G. Tsagarakis, ``A real-time
  identification and tracking method for the musculoskeletal model of human
  arm,'' in \emph{2018 IEEE International Conference on Systems, Man, and
  Cybernetics (SMC)}.\hskip 1em plus 0.5em minus 0.4em\relax IEEE, 2018, pp.
  3472--3479.

\bibitem{durandau2017robust}
G.~Durandau, D.~Farina, and M.~Sartori, ``Robust real-time musculoskeletal
  modeling driven by electromyograms,'' \emph{IEEE transactions on biomedical
  engineering}, vol.~65, no.~3, pp. 556--564, 2017.

\bibitem{hayashibe2011muscle}
M.~Hayashibe, G.~Venture, K.~Ayusawa, and Y.~Nakamura, ``Muscle strength and
  mass distribution identification toward subject-specific musculoskeletal
  modeling,'' in \emph{2011 IEEE/RSJ International Conference on Intelligent
  Robots and Systems}.\hskip 1em plus 0.5em minus 0.4em\relax IEEE, 2011, pp.
  3701--3707.

\bibitem{venture2005identifying}
G.~Venture, K.~Yamane, and Y.~Nakamura, ``Identifying musculo-tendon parameters
  of human body based on the musculo-skeletal dynamics computation and
  hill-stroeve muscle model,'' in \emph{5th IEEE-RAS International Conference
  on Humanoid Robots, 2005.}\hskip 1em plus 0.5em minus 0.4em\relax IEEE, 2005,
  pp. 351--356.

\bibitem{goffe1994global}
W.~L. Goffe, G.~D. Ferrier, and J.~Rogers, ``Global optimization of statistical
  functions with simulated annealing,'' \emph{Journal of econometrics},
  vol.~60, no. 1-2, pp. 65--99, 1994.

\bibitem{rao2006influence}
G.~Rao, D.~Amarantini, E.~Berton, and D.~Favier, ``Influence of body
  segments’ parameters estimation models on inverse dynamics solutions during
  gait,'' \emph{Journal of biomechanics}, vol.~39, no.~8, pp. 1531--1536, 2006.

\bibitem{degrave2019differentiable}
J.~Degrave, M.~Hermans, J.~Dambre, and F.~Wyffels, ``A differentiable physics
  engine for deep learning in robotics,'' \emph{Frontiers in neurorobotics},
  vol.~13, p.~6, 2019.

\bibitem{newbury2024review}
R.~Newbury, J.~Collins, K.~He, J.~Pan, I.~Posner, D.~Howard, and A.~Cosgun, ``A
  review of differentiable simulators,'' \emph{IEEE Access}, 2024.

\bibitem{paszke2017automatic}
A.~Paszke, S.~Gross, S.~Chintala, G.~Chanan, E.~Yang, Z.~DeVito, Z.~Lin,
  A.~Desmaison, L.~Antiga, and A.~Lerer, ``Automatic differentiation in
  pytorch,'' 2017.

\bibitem{sutanto2020encoding}
G.~Sutanto, A.~Wang, Y.~Lin, M.~Mukadam, G.~Sukhatme, A.~Rai, and F.~Meier,
  ``Encoding physical constraints in differentiable newton-euler algorithm,''
  in \emph{Learning for Dynamics and Control}.\hskip 1em plus 0.5em minus
  0.4em\relax PMLR, 2020, pp. 804--813.

\bibitem{le2021differentiable}
Q.~Le~Lidec, I.~Kalevatykh, I.~Laptev, C.~Schmid, and J.~Carpentier,
  ``Differentiable simulation for physical system identification,'' \emph{IEEE
  Robotics and Automation Letters}, vol.~6, no.~2, pp. 3413--3420, 2021.

\bibitem{lutter2021differentiable}
M.~Lutter, J.~Silberbauer, J.~Watson, and J.~Peters, ``Differentiable physics
  models for real-world offline model-based reinforcement learning,'' in
  \emph{2021 IEEE International Conference on Robotics and Automation
  (ICRA)}.\hskip 1em plus 0.5em minus 0.4em\relax IEEE, 2021, pp. 4163--4170.

\bibitem{granados2022model}
E.~Granados, A.~Boularias, K.~Bekris, and M.~Aanjaneya, ``Model identification
  and control of a low-cost mobile robot with omnidirectional wheels using
  differentiable physics,'' in \emph{2022 International Conference on Robotics
  and Automation (ICRA)}.\hskip 1em plus 0.5em minus 0.4em\relax IEEE, 2022,
  pp. 1358--1364.

\bibitem{wang2023real2sim2real}
K.~Wang, W.~R. Johnson, S.~Lu, X.~Huang, J.~Booth, R.~Kramer-Bottiglio,
  M.~Aanjaneya, and K.~Bekris, ``Real2sim2real transfer for control of
  cable-driven robots via a differentiable physics engine,'' in \emph{2023
  IEEE/RSJ International Conference on Intelligent Robots and Systems
  (IROS)}.\hskip 1em plus 0.5em minus 0.4em\relax IEEE, 2023, pp. 2534--2541.

\bibitem{amos2017optnet}
B.~Amos and J.~Z. Kolter, ``Optnet: Differentiable optimization as a layer in
  neural networks,'' in \emph{International conference on machine
  learning}.\hskip 1em plus 0.5em minus 0.4em\relax PMLR, 2017, pp. 136--145.

\bibitem{heiden2020augmenting}
E.~Heiden, D.~Millard, E.~Coumans, and G.~S. Sukhatme, ``Augmenting
  differentiable simulators with neural networks to close the sim2real gap,''
  \emph{arXiv preprint arXiv:2007.06045}, 2020.

\bibitem{wang2023pypose}
C.~Wang, D.~Gao, K.~Xu, J.~Geng, Y.~Hu, Y.~Qiu, B.~Li, F.~Yang, B.~Moon,
  A.~Pandey \emph{et~al.}, ``Pypose: A library for robot learning with
  physics-based optimization,'' in \emph{Proceedings of the IEEE/CVF Conference
  on Computer Vision and Pattern Recognition}, 2023, pp. 22\,024--22\,034.

\bibitem{zajac1989muscle}
F.~E. Zajac, ``Muscle and tendon: properties, models, scaling, and application
  to biomechanics and motor control.'' \emph{Critical reviews in biomedical
  engineering}, vol.~17, no.~4, pp. 359--411, 1989.

\bibitem{zuo2024tsinghua}
C.~Zuo, K.~He, J.~Shao, and Y.~Sui, ``Self model for embodied intelligence:
  Modeling full-body human musculoskeletal system and locomotion control with
  hierarchical low-dimensional representation,'' in \emph{2024 IEEE
  International Conference on Robotics and Automation (ICRA)}.\hskip 1em plus
  0.5em minus 0.4em\relax IEEE, 2024, pp. 13\,062--13\,069.

\bibitem{roy2014bodydyn}
R.~Featherstone, \emph{Rigid body dynamics algorithms}.\hskip 1em plus 0.5em
  minus 0.4em\relax Springer, 2014.

\bibitem{rika2022diffcloud}
P.~Sundaresan, R.~Antonova, and J.~Bohg, ``Diffcloud: Real-to-sim from point
  clouds with differentiable simulation and rendering of deformable objects,''
  in \emph{2022 IEEE/RSJ International Conference on Intelligent Robots and
  Systems (IROS)}.\hskip 1em plus 0.5em minus 0.4em\relax IEEE, 2022, pp.
  10\,828--10\,835.

\bibitem{kingma2014adam}
D.~P. Kingma and J.~Ba, ``Adam: A method for stochastic optimization,''
  \emph{arXiv preprint arXiv:1412.6980}, 2014.

\bibitem{gaz2019franka}
C.~Gaz, M.~Cognetti, A.~Oliva, P.~Giordano, and A.~D. Luca., ``Dynamic
  identification of the franka emika panda robot with retrieval of feasible
  parameters using penalty-based optimization.'' \emph{IEEE Robotics and
  Automation Letters}, vol.~4, no.~4, pp. 4147--4154, 2019.

\bibitem{gray2021minimize}
G.~Gray, ``Pytorch minimize,'' \url{https://github.com/gngdb/pytorch-minimize},
  2021.

\bibitem{leva1996ajustments}
P.~D. Leva, ``Adjustments to zatsiorsky-seluyanov's segment inertia
  parameters.'' \emph{Journal of biomechanics}, vol.~29, no.~9, pp. 1223--1230,
  1996.

\bibitem{gaz2019dynamic}
C.~Gaz, M.~Cognetti, A.~Oliva, P.~R. Giordano, and A.~De~Luca, ``Dynamic
  identification of the franka emika panda robot with retrieval of feasible
  parameters using penalty-based optimization,'' \emph{IEEE Robotics and
  Automation Letters}, vol.~4, no.~4, pp. 4147--4154, 2019.

\bibitem{atkeson1986estimation}
C.~G. Atkeson, C.~H. An, and J.~M. Hollerbach, ``Estimation of inertial
  parameters of manipulator loads and links,'' \emph{The International Journal
  of Robotics Research}, vol.~5, no.~3, pp. 101--119, 1986.

\end{thebibliography}

\end{document}